\pdfoutput=1

\documentclass[11pt]{article}

\usepackage{ACL2023}

\usepackage{times}
\usepackage{latexsym}
\usepackage{multirow}
\usepackage{tabularx}

\usepackage{graphicx}
\usepackage{makecell}

\usepackage{algorithm}

\usepackage{microtype}
\usepackage{algorithmic}
\usepackage{makecell}
\usepackage{cjhebrew}
\usepackage{amsmath}
\usepackage{graphicx}
\usepackage{url}
\usepackage{multirow}
\usepackage{comment}
\usepackage{amssymb}
\usepackage{balance}
\usepackage{mathtools}
\usepackage{xcolor}
\usepackage{colortbl}
\usepackage{bbm}
\usepackage{enumitem}
\usepackage{marvosym}
\usepackage{textcomp}
\usepackage{array}
\usepackage{booktabs}
\usepackage{tabularx}

\usepackage[T1]{fontenc}

\usepackage[utf8]{inputenc}

\usepackage{microtype}

\usepackage{inconsolata}

\title{A comprehensive evaluation of ChatGPT's zero-shot Text-to-SQL capability}

  \author{Aiwei Liu$^{1}$, Xuming Hu$^{1}$, Lijie Wen$^{1}$, Philip S. Yu$^{1,2}$\\
  $^1$Tsinghua University\\  $^2$University of Illinois at Chicago\\
  $^1$\texttt{\{liuaw20, hxm19\}@mails.tsinghua.edu.cn}\\
  $^1$\texttt{wenlj@tsinghua.edu.cn}
  $^2$\texttt{psyu@uic.edu}\\
  }

\begin{document}
\maketitle

\begin{abstract}

This paper presents the first comprehensive analysis of ChatGPT's Text-to-SQL ability. Given the recent emergence of large-scale conversational language model ChatGPT and its impressive capabilities in both conversational abilities and code generation, we sought to evaluate its Text-to-SQL performance.
We conducted experiments on 12 benchmark datasets with different languages, settings, or scenarios, and the results demonstrate that ChatGPT has strong text-to-SQL abilities. Although there is still a gap from the current state-of-the-art (SOTA) model performance, considering that the experiment was conducted in a zero-shot scenario, ChatGPT's performance is still impressive. Notably, in the ADVETA (RPL) scenario, the zero-shot ChatGPT even outperforms the SOTA model that requires fine-tuning on the Spider dataset by 4.1\%, demonstrating its potential for use in practical applications. To support further research in related fields, we have made the data generated by ChatGPT publicly available at \href{https://github.com/THU-BPM/chatgpt-sql}{https://github.com/THU-BPM/chatgpt-sql}.
\end{abstract}

\section{Introduction}

With the increasing attention given to large-scale language models, they have become an essential component in natural language processing. As the size of pre-trained models grows, their usage is also gradually changing. Different from models such as BERT \cite{devlin2018bert} and T5 \cite{raffel2020exploring}, which require fine-tuning with a small amount of data, models such as GPT-3 \cite{brown2020language}, require the prompt design to generate target outputs. The recent ChatGPT\footnote{https://chat.openai.com/} model, which employs Reinforcement Learning for Human Feedback (RLHF) \cite{christiano2017deep}, simplifies prompt design, enabling better utilization of the zero-shot ability of large-scale pre-trained models in a conversational way. Based on this, many works have begun to analyze the zero-shot ability of ChatGPT in various natural language processing tasks, such as information extraction \cite{wei2023zero}, text summarization \cite{wang2023cross}, and mathematical abilities \cite{frieder2023mathematical}. Due to ChatGPT's strong ability in code generation and the fact that code generation models usually require a large amount of annotated data to produce good results, a zero-shot code generation model is very important. This paper first conducts a comprehensive evaluation of ChatGPT's zero-shot performance on a challenging code generation task: Text-to-SQL.

The Text-to-SQL task involves converting user input text into SQL statements that can be executed on a database, allowing non-expert users to better access the contents of a database. The design of Text-to-SQL models is typically challenging because they need to work across different databases and consider various user text input text and database structures. Due to the complexity of the Text-to-SQL task, a comprehensive evaluation of its performance requires consideration of a variety of scenarios in addition to the classic Spider dataset \cite{yu2018spider}. For example, Spider-SYN \cite{gan2021towards} focuses on scenarios where the data schema mentioned in the user text input is synonymous with the database schema, Spider-DK \cite{gan2021exploring}  considers scenarios where the input question contains additional knowledge, Spider-CG \cite{gan-etal-2022-measuring-and} emphasizes the combination generalization ability of models, and ADVETA \cite{pi-etal-2022-towards} considers scenarios where column names in the database have been modified. Additionally, to better reflect real-world scenarios, SParC\cite{yu2019sparc} and CoSQL \cite{yu2019cosql} incorporate multi-turn interaction between the user and the system. Finally, to evaluate models' multilingual capabilities, CSpider \cite{min2019pilot} and DuSQL \cite{wang2020dusql} evaluate Text-to-SQL performance in Chinese.

During our experiments, we evaluate the ability of ChatGPT on 12 different Text-to-SQL benchmark datasets. Based on the experimental results, we conclude the following observations:
\begin{enumerate}
    \item Compared to the current state-of-the-art (SOTA) model that uses complete training data, ChatGPT without using task-specific training data only performs 14\% worse. This already demonstrates that ChatGPT is a strong zero-shot Text-to-SQL converter.
    \item The robustness of ChatGPT in generating SQL statements is very strong, and the performance gap between ChatGPT and the SOTA models is only 7.8\% on some robustness settings of the Spider dataset, which is lower than the 14\% gap on the standard Spider dataset.
    \item  In the ADVETA \cite{pi-etal-2022-towards} scenario where the column names in the database are adversarially modified, ChatGPT's performance even surpasses that of the current SOTA models by 4.1\%.
    \item The Exact Match metric of the data generated by ChatGPT is very low because there are many different ways to express SQLs with the same purpose. Therefore, we mainly use execution accuracy as the evaluation metric.
\end{enumerate}
Overall, our experiments demonstrate that ChatGPT has strong Text-to-SQL capabilities and robustness, and it outperforms SOTA models in certain scenarios.

\section{Method}

\begin{figure}
  \includegraphics[width=0.48\textwidth]{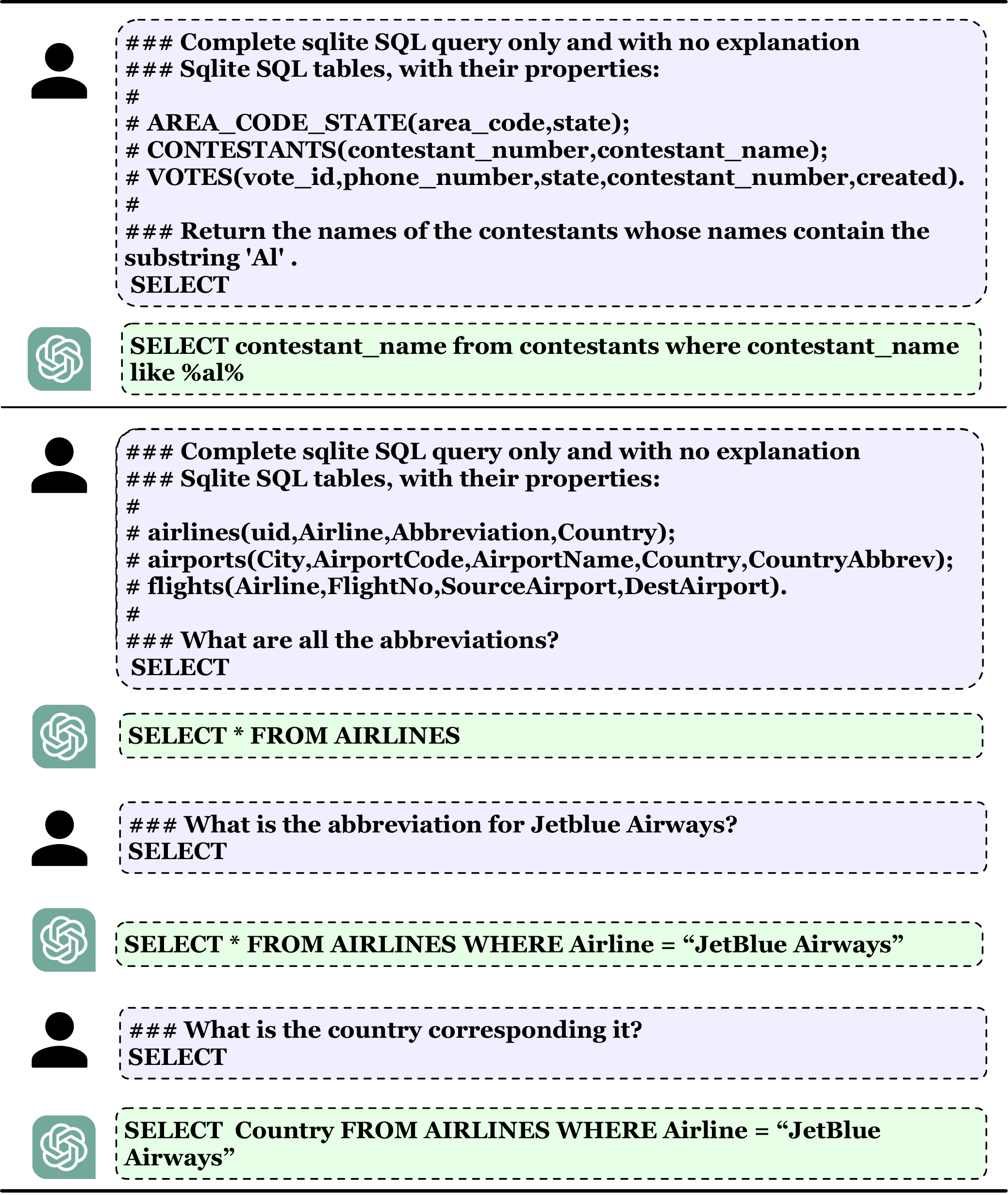}
  \caption{Example prompts for Text-to-SQL using ChatGPT. The prompt at the top is for a single-turn scenario, while the one below is for multi-turn scenarios where only new questions are added in each interaction.}
  \label{fig:intro}
\end{figure}

In order to enable ChatGPT to generate accurate SQL outputs, we utilized the prompt as shown in Figure \ref{fig:intro}. To ensure a fair demonstration of ChatGPT's Text-to-SQL capabilities,  we directly adopted the Text-to-SQL prompt used in the OpenAI demo webwite\footnote{https://platform.openai.com/examples/default-sql-translate} without conducting further prompt exploration.

The upper half of Figure \ref{fig:intro} represents the prompt in a single-turn Text-to-SQL scenario, where only the database and question information is required in the prompt. Meanwhile, in order to facilitate further evaluations, we emphasize in the prompt that the generated SQL statements can be executed in an SQLite database. The lower half of Figure \ref{fig:intro} represents the prompt in a multi-turn Text-to-SQL scenario, where the prompt for the first interaction is the same as that in the single-turn scenario, and for subsequent interactions, only the new questions are required.

\section{Experiment}

\begin{table*}[bt!]
\centering
\resizebox{\linewidth}{!}{
\begin{tabular}{llccccccccc}
\toprule
\multicolumn{1}{c}{\multirow{2}{*}{Methods / Datasets}}  & \multicolumn{3}{c}{\textsc{Spider}} & \multicolumn{3}{c}{\textsc{Spider-SYN}} & \multicolumn{3}{c}{\textsc{Spider-Realistic}}  \\
\cmidrule(lr){2-4} \cmidrule(lr){5-7} \cmidrule(lr){8-10} 
& \textbf{VA}  & \textbf{EX} & \textbf{TS}& \textbf{VA}  & \textbf{EX} & \textbf{TS} & \textbf{VA}  & \textbf{EX} & \textbf{TS}  \\
\hline
T5-3B + PICARD  & 98.4 &  79.3 & 69.4  & 98.2 & 69.8 & 61.8 & 97.1 & 71.4 & 61.7\\
RASAT + PICARD  & 98.8& 80.5 & 70.3 & 98.3 & 70.7 & 62.4 & 97.4 & 71.9 & 62.6\\
RESDSQL-3B + NatSQL & 99.1 & 84.1 & 73.5 & 98.8 & 76.9 & 66.8 & 98.4 & 81.9 & 70.1  \\
ChatGPT & 97.7 & 70.1(14$\downarrow$)  &  60.1 & 96.2 & 58.6(18.3$\downarrow$) & 48.5 & 96.8 & 63.4(18.5 $\downarrow$) & 49.2\\
\bottomrule
\end{tabular}}
\caption{Comparison of the performance of ChatGPT and other models on Spider, Spider-SYN, and Spider-Realistic datasets.}
\label{tab:spider-performance}
\end{table*}
\begin{table*}[bt!]
\centering
\resizebox{\linewidth}{!}{
\begin{tabular}{llccccccccccccccccc}
\toprule
\multicolumn{1}{c}{\multirow{2}{*}{Methods / Datasets}}  & \multicolumn{3}{c}{\textsc{Spider-DK}} & \multicolumn{3}{c}{\textsc{ADVETA(rpl)}}  & \multicolumn{3}{c}{\textsc{ADVETA(add)}} \\
\cmidrule(lr){2-4}  \cmidrule(lr){5-7}   \cmidrule(lr){8-10}  \cmidrule(lr){11-13}
& \textbf{VA}  & \textbf{EX} & \textbf{TS}  &  \textbf{VA}  & \textbf{EX} & \textbf{TS} & \textbf{VA}  & \textbf{EX} & \textbf{TS}\\ 
\hline
T5-3B + PICARD  & 97.8 & 62.5 & -  & 92.7  & 50.6 & -  & 97.2 & 69.4 & -  \\
RASAT + PICARD  & 98.5 & 63.9 & - & 92.9  & 51.5 & - & 97.4 & 70.7 & -  \\
RESDSQL-3B + NatSQL & 98.8 & 66.0 & -  & 93.9 & 54.4 & - & 97.9 & 71.9 & -   \\
ChatGPT & 96.4 & 62.6(3.4 $\downarrow$) & - & 91.4 & 58.5(\textbf{4.1} $\uparrow$) & - & 93.1 & 68.1(3.8 $\downarrow$) & - \\
\bottomrule
\end{tabular}}
\caption{Performance of different methods on the Spider-DK, ADVETA(RPL) and ADVETA(ADD) benchmark datasets.}
\label{tab:spider-dk}
\end{table*}
\subsection{Experiment Setup}

\noindent\textbf{Datasets.} 
 We conduct extensive experiments on twelve public benchmark datasets as follows: (1) \textbf{Spider} \cite{yu2018spider} is a large-scale cross-domain Text-to-SQL benchmark. It contains 8659 training samples across 146 databases and 1034 evaluation samples across 20 databases.  (2) \textbf{Spider-SYN} \cite{gan2021towards} is a challenging variant of the Spider evaluation dataset. Spider-SYN is constructed by manually modifying natural language questions with synonym substitutions.  (3) \textbf{Spider-DK} \cite{gan2021exploring}  is a human-curated dataset based on Spider, which samples 535 question-SQL pairs across 10 databases from the Spider development set and modifies them to incorporate the domain knowledge. (4) \textbf{Spider-Realistic} \cite{deng2020structure} is a new evaluation set based on the Spider dev set with explicit mentions of column names removed, which contains 508 samples. (5) \textbf{Spider-CG(SUB)} and \textbf{Spider-CG(APP)} \cite{gan-etal-2022-measuring-and} are two evaluation datasets to measure the compositional generalization of models, which is constructed by sub-sentence substitution between different
examples and appending a sub-sentence into another sentence separately. (6) \textbf{ADVETA(rpl)} and \textbf{ADVETA(add)} \cite{pi-etal-2022-towards}  are two challenging test datasets for the Spider dataset which are composed of adversarial replacements of column names and the addition of new column names, respectively. (7) 
 \textbf{CSpider} \cite{min2019pilot} dataset is constructed by translating Spider into Chinese, which is the same size as the origin Spider dataset (8) \textbf{DuSQL} \cite{wang2020dusql} is a larger scale Chinese Text-to-SQL dataset with 23,797 question/SQL pairs. (9) \textbf{SParC} \cite{yu2019sparc} and \textbf{CoSQL} \cite{yu2019cosql} are two multi-turn Text-to-SQL dataset with 1625 and 1007 questions in the dev set separately. \\

\noindent\textbf{Evaluation Metrics.}  We mainly adopt three evaluation metrics which are valid SQL (VA), execution accuracy(EX), and test-suite accuracy (TS). Valid SQL (VA) is the proportion of SQL statements that can be executed successfully. Execution accuracy (EX) is the proportion of data where the execution results match the standard SQL statements. Test-suite accuracy (TS) \cite{zhong2020semantic} could achieve high code coverage from a distilled test suite of the database, which is also based on execution. Note that we do not use  the main-stream exact match accuracy, as SQL queries that achieve the same goal can often be expressed in different ways, making it difficult for zero-shot ChatGPT models to achieve high exact match accuracy. \\

 \begin{table*}[bt!]
\centering
\resizebox{0.8\linewidth}{!}{
\begin{tabular}{llcccccccccccccc}
\toprule
\multicolumn{1}{c}{\multirow{2}{*}{Methods / Datasets}}  & \multicolumn{3}{c}{\textsc{Spider-CG(SUB)}} & \multicolumn{3}{c}{\textsc{Spider-CG(APP)}}  \\
\cmidrule(lr){2-4}  \cmidrule(lr){5-7} 
& \textbf{VA}  & \textbf{EX} & \textbf{TS}  &  \textbf{VA}  & \textbf{EX} & \textbf{TS}\\ 
\hline
T5-3B + PICARD  & 98.4 & 82.1 & 74.3  & 95.8  & 68.0 & 60.5   \\
RASAT + PICARD  & 99.0 & 82.6 & 76.1 & 96.2  & 68.6 & 61.0  \\
RESDSQL-3B + NatSQL & 99.4 & 83.3 & 77.5  & 96.4 & 69.4 & 62.4  \\
ChatGPT & 98.3 & 76.6(6.7 $\downarrow$) & 67.2 & 91.2 & 61.3(8.1 $\downarrow$) & 47.9  \\
\bottomrule
\end{tabular}}
\caption{Performance of different methods on the Spider-CG(SUB) and Spider-CG(APP) benchmark datasets.}
\label{tab:spider-cg}
\end{table*}

\noindent\textbf{Baselines.}  Due to our exclusive reliance on execution-based evaluation, we did not employ baselines such as RatSQL \cite{wang2019rat} and LGESQL \cite{cao2021lgesql}, which generate only SQL skeletons without generating values. Instead, we primarily utilized three baselines: (1) PICARD \cite{scholak2021picard} is a method for constraining auto-regressive decoders of language models through incremental parsing. (2) RASAT \cite{qi2022rasat} introduces relation-aware self-attention into transformer models and also utilizes constrained auto-regressive decoders. (3) RESDSQL \cite{li2023decoupling} proposes a ranking-enhanced encoding and skeleton-aware decoding framework to decouple the schema linking and the skeleton parsing.  Among those, PICARD and RASAT are based on T5-3B \cite{raffel2020exploring} model.
 
\subsection{Main Experiment}

\noindent\textbf{Evaluation on Spider Dataset.} In Table \ref{tab:spider-performance}, we present a comparison between ChatGPT and the current state-of-the-art (SOTA) models. Overall, ChatGPT exhibits a strong Text-to-SQL ability.Despite the 14\% gap in execution accuracy compared to the current SOTA models and a 13.4\% gap in test suite accuracy, it is remarkable that ChatGPT achieved such results in a zero-shot scenario considering that it was not fine-tuned on the Spider training set.\\

\noindent\textbf{Evaluation on Spider-SYN and Spider-Realistic Datasets.} Table \ref{tab:spider-performance} also includes a comparison of ChatGPT's performance on the Spider-SYN and Spider-Realistic datasets.  The main difference between these datasets and the Spider dev set is that they eliminate the explicit appearance of the database schema in the questions. Overall, although ChatGPT still performs well on these two settings, the performance gap between ChatGPT and the original SOTA models becomes slightly larger than that on the Spider dataset. This suggests that the current models have already achieved sufficient robustness in these two scenarios. \\

\noindent\textbf{Evaluation on Spider-DK and ADVETA Datasets.} In Table \ref{tab:spider-dk}, we further compare and analyze ChatGPT's performance on Spider-DK, ADVETA (RPL), and ADVETA (ADD). We find that ChatGPT performs exceptionally well on these datasets, with very small performance gaps compared to the current SOTA models. In fact, ChatGPT outperforms all current SOTA models on ADVETA (RPL). For the Spider-DK dataset, we speculate that ChatGPT's excellent performance is due to its additional knowledge provided by the large-scale pretraining. As for scenarios such as ADVETA, where the dataset's column names undergo adversarial modifications, the poor generalization performance of current models may be due to the significant distribution difference from the original dataset. Overall, ChatGPT exhibits strong robustness in scenarios that require additional knowledge or adversarial modifications are applied to the database column names.  \\

 \begin{table}[bt!]
\centering
\resizebox{\linewidth}{!}{
\begin{tabular}{llccccccccc}
\toprule
\multicolumn{1}{c}{\multirow{2}{*}{Methods / Datasets}}  & \multicolumn{2}{c}{\textsc{SParC}} & \multicolumn{2}{c}{\textsc{CoSQL}}  \\
\cmidrule(lr){2-3} \cmidrule(lr){4-5}
& \textbf{VA}  & \textbf{EX} & \textbf{VA}  & \textbf{EX}   \\
\hline
T5-3B + PICARD  & - &  - & 97.5 & 64.7 \\
RASAT + PICARD  & 98.4& 74.0 & 97.8 & 66.3\\
ChatGPT & 97.3 & 63.1 & 95.8 & 60.7 \\
\bottomrule
\end{tabular}}
\caption{The performance of ChatGPT on two multi-turn Text-to-SQL datasets: SParC and CoSQL.}
\label{tab:multi}
\end{table}

\noindent\textbf{Evaluation on Spider-CG Dataset.} In Table \ref{tab:spider-cg}, we further analyze ChatGPT's ability in the compositional generalization scenario. We found that in Spider-CG (SUB), SQL substructures are replaced to form combinations that do not exist in the training set. In this scenario, ChatGPT even outperforms the original Spider dev set. Even on the more challenging Spider-CG (APP) dataset, ChatGPT achieves strong performance, and the performance gap with SOTA models is relatively smaller than that on the original Spider dataset. Overall, since ChatGPT is a zero-shot model, it is not as affected by compositional generalization as the SOTA models. Overall, zero-shot models have greater advantages in the compositional generalization setting. \\

\begin{table}[bt!]
\centering
\resizebox{\linewidth}{!}{
\begin{tabular}{llccccccccc}
\toprule
\multicolumn{1}{c}{\multirow{2}{*}{Methods / Datasets}}  & \multicolumn{2}{c}{\textsc{CSpider}} & \multicolumn{2}{c}{\textsc{DuSQL}}  \\
\cmidrule(lr){2-3} \cmidrule(lr){4-5}
& \textbf{VA}  & \textbf{EX} & \textbf{VA}  & \textbf{EX}   \\
\hline
ChatGPT & 96.0 & 65.1 & 82.7 & 53.7 \\
\bottomrule
\end{tabular}}
\caption{The performance of ChatGPT on two Chinese Text-to-SQL datasets: CSpider and DuSQL.}
\label{tab:chinese}
\end{table}

\noindent\textbf{Evaluation on multi-turn Text-to-SQL scenarios.} Given ChatGPT's strong contextual modeling ability, we further evaluate its performance on multi-turn Text-to-SQL scenarios: \textsc{SParC} and \textsc{CoSQL}. As shown in Table \ref{tab:multi}, ChatGPT exhibits strong multi-turn Text-to-SQL ability. Although there is still a gap compared to the current SOTA models, the gap is relatively smaller compared to the single-turn Spider dataset. Meanwhile, ChatGPT also performs better on CoSQL datasets with more average interactions, which also indicates that ChatGPT's strong contextual modeling ability is very helpful for multi-turn Text-to-SQL. \\

\noindent\textbf{Evaluation on Chinese Text-to-SQL scenarios.} We further evaluate ChatGPT's Text-to-SQL ability on other languages in Table \ref{tab:chinese}. The experiments are mainly conducted on two datasets, CSpider and DuSQL, where only the questions are in Chinese for CSpider and both the schema names and questions are in Chinese for DuSQL. The results show that while ChatGPT performs well in the Chinese Text-to-SQL scenario, there is still a performance gap compared to the English Text-to-SQL scenario. Moreover, the performance is even worse when the table names and column names are also in Chinese, with a large number of generated SQL queries being non-executable and a lower execution accuracy. This suggests the cross-lingual generalization ability of ChatGPT requires further improvement.

\subsection{Case Study}

\renewcommand{\arraystretch}{1.3}
\begin{table}[t!]
\centering
\resizebox{1.00\linewidth}{!}{
\begin{tabular}{l}
\toprule
\begin{tabular}{l@{}@{}} 
\multicolumn{1}{@{}p{1.1\linewidth}}{\textbf{Question:} Show the stadium name and the number of concerts in each stadium.}
\\
\multicolumn{1}{@{}p{1.1\linewidth}}{\textbf{ChatGPT:} \texttt{SELECT stadium.Name, COUNT(concert.concert\_ID) FROM stadium  \textbf{{\color{red} LEFT JOIN }} concert ON stadium.Stadium\_ID = concert.Stadium\_ID  GROUP BY stadium.Name; }}
\\ 
\multicolumn{1}{@{}p{1.1\linewidth}}{\textbf{Gold:} \texttt{SELECT T2.name ,  count(*) FROM concert AS T1 \textbf{{\color{blue} JOIN}} stadium AS T2 ON T1.stadium\_id  =  T2.stadium\_id GROUP BY T1.stadium\_id }}
\end{tabular}     
\\ 
\hline
\begin{tabular}[c]{@{}l@{}}
\multicolumn{1}{@{}p{1.1\linewidth}}{\textbf{Question:}How many car models were produced by the maker with full name American Motor Company?}
\\
\multicolumn{1}{@{}p{1.1\linewidth}}{\textbf{ChatGPT:} \texttt{SELECT COUNT(*)  FROM model\_list  WHERE  \textbf{{\color{red} Maker }} = 'American Motor Company'}} 
\\
\multicolumn{1}{@{}p{1.1\linewidth}}{\textbf{Gold:} \texttt{SELECT count(*) FROM CAR\_MAKERS AS T1 JOIN model\_list AS T2 ON T1.Id  =  T2.Maker WHERE   \textbf{{\color{blue} T1.FullName }}  =  'American Motor Company';}} 
\end{tabular}    
\\
\hline
\begin{tabular}[c]{@{}l@{}}
\multicolumn{1}{@{}p{1.1\linewidth}}{\textbf{Question:} How many cars have a larger accelerate than the car with the largest horsepower?}
\\
\multicolumn{1}{@{}p{1.1\linewidth}}{\textbf{ChatGPT:} \texttt{SELECT COUNT(*) FROM cars\_data WHERE Accelerate > (SELECT \textbf{{\color{red} MAX(Horsepower)}}  FROM cars\_data)}} 
\\
\multicolumn{1}{@{}p{1.1\linewidth}}{\textbf{Gold:} \texttt{SELECT COUNT(*) FROM CARS\_DATA WHERE Accelerate  >  (SELECT \textbf{{\color{blue} Accelerate}}  FROM CARS\_DATA ORDER BY Horsepower DESC LIMIT 1);}} 
\end{tabular}    
\\
\hline
\begin{tabular}[c]{@{}l@{}}
\multicolumn{1}{@{}p{1.1\linewidth}}{\textbf{Question:} What is the abbreviation of Airline  "JetBlue Airways"?}
\\
\multicolumn{1}{@{}p{1.1\linewidth}}{\textbf{ChatGPT:} \texttt{SELECT Abbreviation  FROM airlines  WHERE Airline = \textbf{{\color{red} 'Jetblue Airways'}} ;}} 
\\
\multicolumn{1}{@{}p{1.1\linewidth}}{\textbf{Gold:} \texttt{SELECT Abbreviation FROM AIRLINES WHERE Airline = \textbf{{\color{blue}  "JetBlue Airways"}};}} 
\end{tabular}    
\\
\bottomrule
\end{tabular}
}
\caption{Case study: We selected four cases of incorrect predictions generated by ChatGPT on the Spider development set for analysis.
}\label{tab:example}
\end{table}

In Table 6, we present four typical prediction errors made by ChatGPT on the Spider dev dataset. The first error case shows that ChatGPT tends to design JOIN statements more finely by using LEFT JOIN, but this level of granularity is not present in the original Spider dev dataset. The second error case arises from ChatGPT's confusion regarding the database structure, and it is not clear which column the term "full name" specifically refers to. The third example's error was due to the generated SQL statement lacking correct semantic interpretation, resulting in incorrect output for the "where" clauses with nested SQL statements.  The fourth case of error is due to errors in copying specific values, where the case sensitivity of the original value was not preserved when regenerating the value.

In summary, ChatGPT's errors mostly occur in small details, and some of these issues can be addressed and improved in later stages of development, such as in the first, third, and fourth cases. However, for errors like the second case, which indicate a lack of understanding of the database schema, further improvements to the model's ability may be necessary to resolve them.

\section{Related Work}

Text-to-SQL is an important semantic parsing task that converts natural language questions posed by users into SQL statements that can be executed on a database. On the classic Spider dataset \cite{yu2018spider}, many classic works such as RatSQL \cite{wang2019rat} and LGESQL \cite{cao2021lgesql} have achieved excellent results. Since Text-to-SQL is a very complex task involving both user input questions and database structure, the robustness of the model is crucial. To further explore this issue, \citet{gan2021towards} proposed the Spider-SYN dataset to evaluate the robustness of models under synonym substitution scenarios. Some works, such as Proton \cite{wang2022proton} and ISESL-SQL \cite{liu2022semantic}, are also devoted to improving the robustness of models in this scenario. Meanwhile, many works explore the robustness of the Text-to-SQL task in other scenarios. The Spider-DK dataset \cite{gan2021exploring} evaluates the robustness of models in scenarios requiring additional knowledge. The Spider-Realistic dataset \cite{deng2020structure} removes the explicit appearance of dataset schema information in user questions, thereby increasing the difficulty of the original task. The Spider-CG dataset \cite{gan-etal-2022-measuring-and} evaluates the robustness of models in compositional generalization scenarios. The ADVETA dataset \cite{pi-etal-2022-towards} evaluates the robustness of models in scenarios involving adversarial modifications of database table information. In addition, to verify the robustness of models in cross-lingual scenarios, CSpider \cite{min2019pilot} and DuSQL \cite{wang2020dusql} have been proposed to evaluate the robustness of models in the Chinese language. To evaluate the performance of Text-to-SQL in more realistic scenarios, SParC \cite{yu2019sparc} and CoSQL \cite{yu2019cosql} have been proposed to evaluate the performance of multi-turn Text-to-SQL. Models such as \textsc{STaR} \cite{cai2022star} and CQR-SQL \cite{xiao2022cqr} have also achieved good results in this scenario.

Currently, several methods have been attempted to explore the improvement of large-scale language models for Text-to-SQL models. The PICARD \cite{scholak2021picard} and  RASAT \cite{qi2022rasat} utilize the T5-3B model, but still require the training data for fine-tuning. \citet{rajkumar2022evaluating} investigated the Text-to-SQL capabilities of the GPT3 model in a zero-shot setting. \citet{cheng2022binding} proposed the BINDER model based on the GPT3 codex, which has similar Text-to-SQL generation capabilities with the need for in-context exemplar annotations. However, these works do not provide a comprehensive evaluation of Text-to-SQL and are limited to a few datasets without other robustness settings. In this work, we are the first to evaluate the comprehensive Text-to-SQL capabilities of ChatGPT.

\section{Conclusion}

In this work, we conducted a comprehensive analysis of ChatGPT's zero-shot ability in Text-to-SQL. We found that even without using any training data, ChatGPT still has strong Text-to-SQL ability, although there is still some gap compared to the current SOTA models. Additionally, ChatGPT demonstrated strong robustness, performing relatively better on most robustness benchmarks and even surpassing the current SOTA models on the ADVETA benchmark. Although this paper has made some findings, we only utilize a common prompt to evaluate ChatGPT's ability. And in future work, better prompts could be designed to explore ChatGPT's Text-to-SQL ability.

\section{Future work}

In future work, we will primarily consider the following two directions to further explore ChatGPT's capabilities in the Text-to-SQL task.
Firstly, we will conduct more interactions with ChatGPT to address the issue of generating non-executable SQL statements. We can design ChatGPT to engage in multi-turn dialogues with the provided database error messages to further ensure the validity of generated SQL statements.
Secondly, we will add more highly correlated in-context examples to the prompt to enhance ChatGPT's ability to generate Text-to-SQL.
\bibliography{custom}
\bibliographystyle{acl_natbib}
\end{document}